\documentclass[runningheads]{llncs}
\usepackage{hyperref}
\usepackage{cite}
\usepackage{multirow}
\usepackage[normalem]{ulem}
\usepackage{makecell}
\usepackage{algorithmic}
\usepackage{algorithm}
\usepackage{amsfonts,amsmath}
\usepackage{marvosym}
\useunder{\uline}{\ul}{}

\usepackage[T1]{fontenc}
\usepackage{graphicx}
\usepackage{color}

\begin{document}
\title{Towards A Flexible Accuracy-Oriented Deep Learning Module Inference Latency Prediction Framework for Adaptive Optimization Algorithms}
\titlerunning{Towards A Flexible DNN Inference Latency Prediction Framework}
%
\author{Jingran Shen\inst{1} \and
Nikos Tziritas\inst{2} \and
Georgios Theodoropoulos\inst{3}\textsuperscript{(\Letter)}}
\authorrunning{J. Shen et al.}
%
\institute{Department of Computer Science and Engineering, Southern University of Science and Technology (SUSTech), Shenzhen,
P.R. China\\ \email{petershen815@126.com} \and
Department of Informatics and Telecommunications, University of Thessaly, Lamia, Greece\\ \email{nitzirit@uth.gr} \and
Research Institute for Trustworthy Autonomous Systems and Department of Computer Science and Engineering, Southern University of Science and Technology (SUSTech), Shenzhen,
P.R. China\\ \email{theogeorgios@gmail.com}}
\maketitle              
\begin{abstract}
With the rapid development of Deep Learning, more and more applications on the cloud and edge tend to utilize large DNN (Deep Neural Network) models for improved task execution efficiency as well as decision-making quality. Due to memory constraints, models are commonly optimized using compression, pruning, and partitioning algorithms to become deployable onto resource-constrained devices. As the conditions in the computational platform change dynamically, the deployed optimization algorithms should accordingly adapt their solutions. To perform frequent evaluations of these solutions in a timely fashion, RMs (Regression Models) are commonly trained to predict the relevant solution quality metrics, such as the resulted DNN module inference latency, which is the focus of this paper. Existing prediction frameworks specify different RM training workflows, but none of them allow flexible configurations of the input parameters (e.g., batch size, device utilization rate) and of the selected RMs for different modules. In this paper, a deep learning module inference latency prediction framework is proposed, which i) hosts a set of customizable input parameters to train multiple different RMs per DNN module (e.g., convolutional layer) with self-generated datasets, and ii) automatically selects a set of trained RMs leading to the highest possible overall prediction accuracy, while keeping the prediction time / space consumption as low as possible. Furthermore, a new RM, namely MEDN (Multi-task Encoder-Decoder Network), is proposed as an alternative solution. Comprehensive experiment results show that MEDN is fast and lightweight, and capable of achieving the highest overall prediction accuracy and R-squared value. The Time/Space-efficient Auto-selection algorithm also manages to improve the overall accuracy by 2.5\% and R-squared by 0.39\%, compared to the MEDN single-selection scheme.

\end{abstract}
\section{Introduction}
\label{sec:intro}
\subsection{Background}

There has been an increasing number of cloud/edge-based applications that utilize DNN (Deep Neural Network) models to improve task execution efficiency as well as decision-making quality. Specifically, large DNN models like GPT-3 \cite{gpt3} and SAM \cite{sam} (Segment Anything Model) are facing extra deployment concerns due to their complex structures and heavy model sizes. The memory deficiency problem becomes even more critical when these models need to be deployed to a resource-constrained edge environment. To this end, a large amount of research on different categories of model optimization algorithms has been conducted through the past few years. These algorithms include i) compressing DNN models through quantization techniques or autoencoders \cite{autosplit, branchygnn}, ii) pruning DNN models by reducing model parameters or by modifying model structures \cite{pteenet, edgent, branchygnn, ddnn}, and iii) partitioning DNN models onto multiple devices \cite{autosplit, modnn, latency_driven_model_placement, edgent, branchygnn, multi_user_dnn_partitioning, ddnn}. Table \ref{tab:model-opt-alg} lists several adaptive model optimization algorithms as examples.

\setlength{\arrayrulewidth}{1pt}
\setlength{\tabcolsep}{1.0pt}
\renewcommand{\arraystretch}{1.2}
\begin{table}
\centering
\caption{Adaptive DNN Model optimization algorithms.}
\label{tab:model-opt-alg}
\begin{tabular}{|c|c|c|c|c|}
\hline
\textbf{Algorithm} & \textbf{Compression} & \textbf{Pruning} & \textbf{Partitioning} & \textbf{Description} \\ 
\hline
Auto-Split \cite{autosplit} & \checkmark &   & \checkmark & \makecell{two-way partitioning with \\ module-wise quantization} \\
\hline
PTEENet \cite{pteenet} &   & \checkmark &   & \makecell{adaptive early-exiting} \\
\hline
MoDNN \cite{modnn} &   &   & \checkmark & \makecell{Linear module \\ weight parititioning} \\
\hline
\makecell{$k$-way DAG \\ Partitioning \cite{latency_driven_model_placement}} &   &   & \checkmark & \makecell{$k$-way partitioning} \\
\hline
Edgent \cite{edgent} &   & \checkmark & \checkmark & \makecell{two-way partitioning \\ and adaptive early-exiting} \\
\hline
\makecell{Branchy \\ GNN \cite{branchygnn}} & \checkmark & \checkmark & \checkmark & \makecell{two-way partitioning, \\ adaptive early-exiting, \\ and intermediate \\ feature compression} \\
\hline
\end{tabular}
\end{table}

Occasionally, the varying available memory and network bandwidth of the devices may affect the quality of the currently provided solution. Hence, model optimization algorithms should adjust their solutions accordingly to adapt to these environment changes, which means that solution quality evaluations will happen frequently. To perform the evaluations in a timely fashion due to real-time requirements, RMs (Regression Models) are commonly trained to predict relevant solution metrics - specifically in this paper, the resulted module inference latency. These trained RMs can also be applied to other adaptive optimization algorithms like inference request batching \cite{batch_model_mgmt, multi_user_batch} and input data partitioning \cite{modnn, edgeflow, coedge}. Several prediction frameworks \cite{edgent, nn_meter, swfu} have been introduced to specify different RM training workflows, but none of them allow flexible configurations of the input parameters (e.g., batch size, FLOPS, device utilization rate) and of the selected RMs for different DNN modules. Since different modules have varying degrees of structure and computation complexity, they require different regression models and input parameters for prediction. Hence, a flexible prediction framework that can accommodate such diversity is essential.
\subsection{Gap Analysis on Related Work}

In this paper, a DNN module inference latency prediction framework is proposed to address the aforementioned problem by i) hosting a set of customizable input parameters to train multiple different RMs per DNN module (e.g., convolutional layer) with self-generated datasets, and ii) automatically selecting a set of trained RMs leading to the highest possible overall prediction accuracy, while keeping the prediction time / space consumption as low as possible. Besides, a new RM named MEDN (Multi-task Encoder-Decoder Network) is proposed as an alternative solution for RM selection.

\begin{figure}
    \centering
    \includegraphics[width=0.85\textwidth]{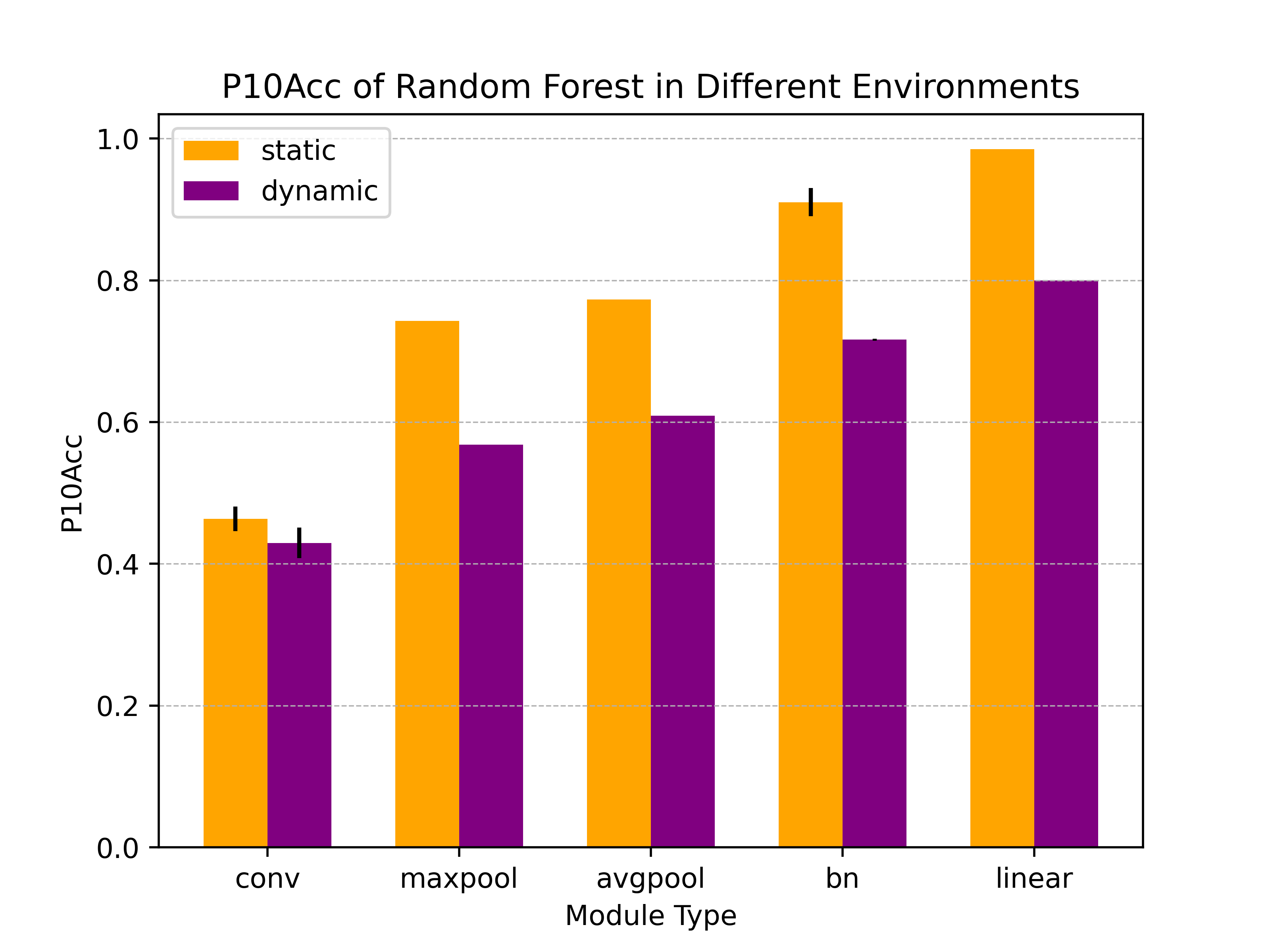}
    \caption{P10-Accuracy of Random Forest in different environments.}
    \label{fig:env}
\end{figure}

Training a universal RM for all modules \cite{stanford_profiler, swfu} is technically more difficult and non-scalable due to the ever-increasing number of deep-learning modules. Instead, the proposed framework focuses on training one RM per module and forming a group of prediction experts through RM auto-selection, while existing frameworks only support the single-selection scheme that uses one RM for all modules. Furthermore, the proposed framework measures the available memory and utilization rate of the device to further improve the prediction accuracy in a dynamic environment, where other workloads are also consuming device resources. Other frameworks like Edgent \cite{edgent} and nn-Meter \cite{nn_meter} do not cover this scenario. As shown in Figure \ref{fig:env}, random forests trained and refined by nn-Meter face significant accuracy drops when switching to a dynamic environment. Additionally, nn-Meter does not handle cases where the data batch size is more than $1$. Finally, the proposed framework introduces Inferable Parameters to calculate more meaningful input features (e.g., data/module size) as extra information for RMs to handle, which is not applied by Edgent \cite{edgent}. In addition, the proposed framework introduces a newly designed encoder-decoder network - MEDN (Multi-task Encoder-Decoder Network), which is faster on predictions and far more light-weight than random forests utilized by nn-Meter. Besides, MEDN better captures features from the input parameters and generally outperforms other Deep RMs (e.g., Multi-Layer Perceptron, namely MLP \cite{stanford_profiler, swfu}) on both accuracy and R-squared metrics. Table \ref{tab:gap} summarizes the features of the proposed framework compared to other relevant latency prediction frameworks.

\setlength{\arrayrulewidth}{1pt}
\setlength{\tabcolsep}{3.0pt}
\renewcommand{\arraystretch}{1.2}
\begin{table}
\centering
\caption{Gap analysis of relevant researches on the proposed topic.}
\label{tab:gap}
\begin{tabular}{|c|c|c|c|c|}
\hline
\textbf{Features} & Edgent \cite{edgent} & nn-Meter \cite{nn_meter} & SWFU \cite{swfu} & \textbf{\makecell{Proposed \\ Framework}} \\ 
\hline
\makecell{Utilize RM} & \checkmark & \checkmark & \checkmark & \checkmark \\
\hline
\makecell{Handle larger batch sizes} & \checkmark &   & \checkmark & \checkmark \\
\hline
\makecell{Infer extra information} &   & \checkmark & \checkmark & \checkmark \\
\hline
\makecell{Measure device dynamics} &   &   & \checkmark & \checkmark \\
\hline
Utilize Deep RM &   &   & \checkmark & \checkmark \\
\hline
Use Encoder-Decoder &   &   &   & \checkmark \\
\hline
Support RM Auto-selection &   &   &   & \checkmark \\
\hline
\end{tabular}
\end{table}

The rest of the paper is structured as follows: Section \ref{sec:intro} discusses the motivation for introducing the proposed framework and briefly states its advantages compared to other existing DNN module inference latency prediction frameworks. Section \ref{sec:framework} outlines the key components and operations of the proposed framework. Section \ref{sec:exp} presents the experimental frame as well as a quantitative analysis and evaluation of the proposed system. Section \ref{sec:conclusion} summarizes the paper and lists potential future research directions.
\section{Framework Design}
\label{sec:framework}

The key objective of the proposed framework is to support flexible RM training configurations as well as accuracy-oriented automatic RM selections. The following components described in this section constitute the ability of the proposed framework to achieve the stated objective.
\subsection{Parameters}

The input parameters are categorized into three genres, namely i) Sampling Parameters, ii) Measurable Parameters, and iii) Inferable Parameters, to help with systematic analysis on the their importance in a dynamic environment, where other workloads on the device affect the module inference performances. 

\setlength{\arrayrulewidth}{1pt}
\setlength{\tabcolsep}{5.0pt}
\renewcommand{\arraystretch}{1.1}
\begin{table}
\centering
\caption{Parameters for each module.}
\label{tab:params}
\begin{tabular}{|c|c|c|c|}
\hline
\textbf{Module} & \textbf{\makecell{Sampling \\ Parameters}} & \textbf{\makecell{Measurable \\ Parameters}} & \textbf{\makecell{Inferable \\ Parameters}} \\ 
\hline
avgpool & \makecell{$N$, $C_i$, $L$, $K$, $S$} & \makecell{$M$, $U$} & $N_d$ \\
bn & \makecell{$N$, $L$, $C_i$} & \makecell{$M$, $U$} & $N_d, N_m$ \\
conv & \makecell{$N$, $L$, $C_i$, $C_o$, $K$, $S$} & \makecell{$M$, $U$} & $N_d, N_m$ \\
linear & \makecell{$N$, $C_i$, $C_o$} & \makecell{$M$, $U$} & $N_d, N_m$ \\
maxpool & \makecell{$N$, $C_i$, $L$, $K$, $S$} & \makecell{$M$, $U$} & $N_d$ \\
\hline
\end{tabular}
\end{table}

The Sampling Parameters refer to the original parameters that can be used to construct the corresponding data and module. As an example, a convolutional module can be constructed by specifying the Sampling Parameters including the input/output channels $C_i, C_o \in [3, 512]$, kernel size $K \in \{ 1, 3, 5, 7, 9 \}$, and stride $S \in \{ 1, 2, 4 \}$. The corresponding input data for the convolutional module can be constructed by specifying the batch size $N \in [1, 64]$, input height/weight $L \in \{ 224, 112, 56, 32, 28, 27, 14, 13, 8, 7 \}$ along with the input channels $C_i$. Apart from these original parameters, the available memory $M$ (in bytes) and utilization rate $U \in [0, 1.0]$ of the device should be recorded as Measurable Parameters to capture the features from the dynamic environment. Intuitively, RMs have to be informed about whether the device is free and available to commit enough resources to module inference tasks. When the device is heavily occupied with other running jobs, RMs should predict a relatively slower inference speed than the normal scenario (i.e., static environment). Finally, Inferable Parameters describe the set of parameters whose values can be calculated from other Sampling/Measurable/Inferable Parameters. This category of parameters generally offers fine-grained information to help RMs better understand the input features. As an example, the number of module weight parameters $N_m$ indicates the module complexity and the input data size $N_d$ reflects the computation overheads. 

Depending on the structure of different RMs, different parameter sets can be configured flexibly during the training procedure. Table \ref{tab:params} summarizes the available parameters for various modules including Convolution, Max/Average Pooling, Batch Normalization (abbreviated as "bn" in this paper), and Linear.
\subsection{Regression Model Training Workflow}

\begin{figure}
    \centering
    \includegraphics[width=0.5\textwidth]{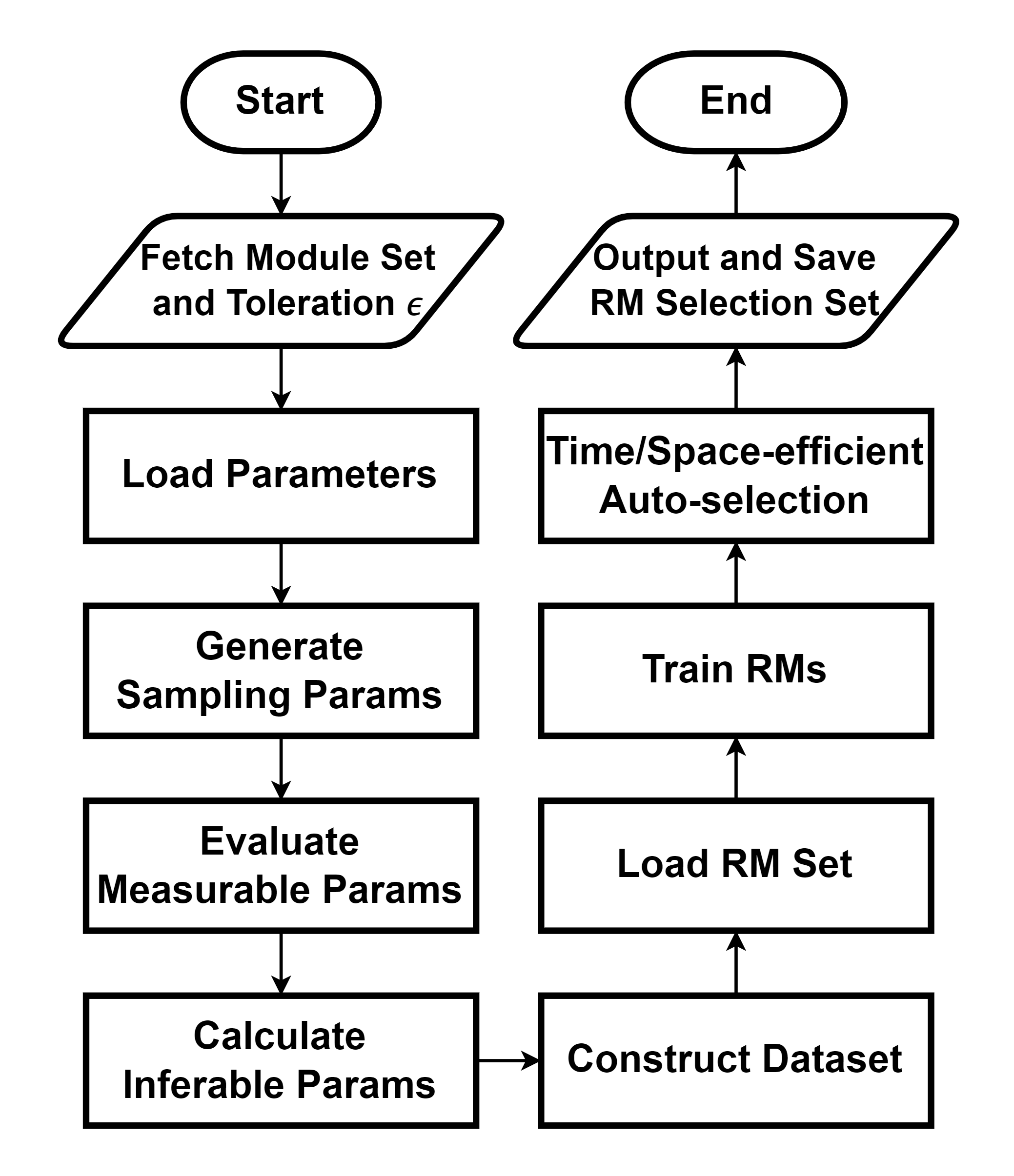}
    \caption{RM training workflow of the framework.}
    \label{fig:workflow}
\end{figure}

Figure \ref{fig:workflow} demonstrates the RM training workflow of the proposed framework. Basically, training data samples are constructed by packing all configured parameter values as input features. In detail, Sampling Parameter values are randomly generated in the specified ranges and then used to construct the corresponding data as well as modules for ground-truth latency measurement. Next, device status is evaluated into Measurable Parameters and all Inferable Parameters are calculated from the existing parameters. Finally, the packed parameter values and the measured latency results are combined into a dataset, which is then used to train a set of pre-configured RMs. As the last step of the training workflow, an auto-selection algorithm outputs a set of RMs achieving the highest possible overall prediction accuracy, while keeping the prediction time / space consumption as low as possible.

\subsection{Towards a New Regression Model}

\begin{figure}
    \centering
    \includegraphics[width=0.9\textwidth]{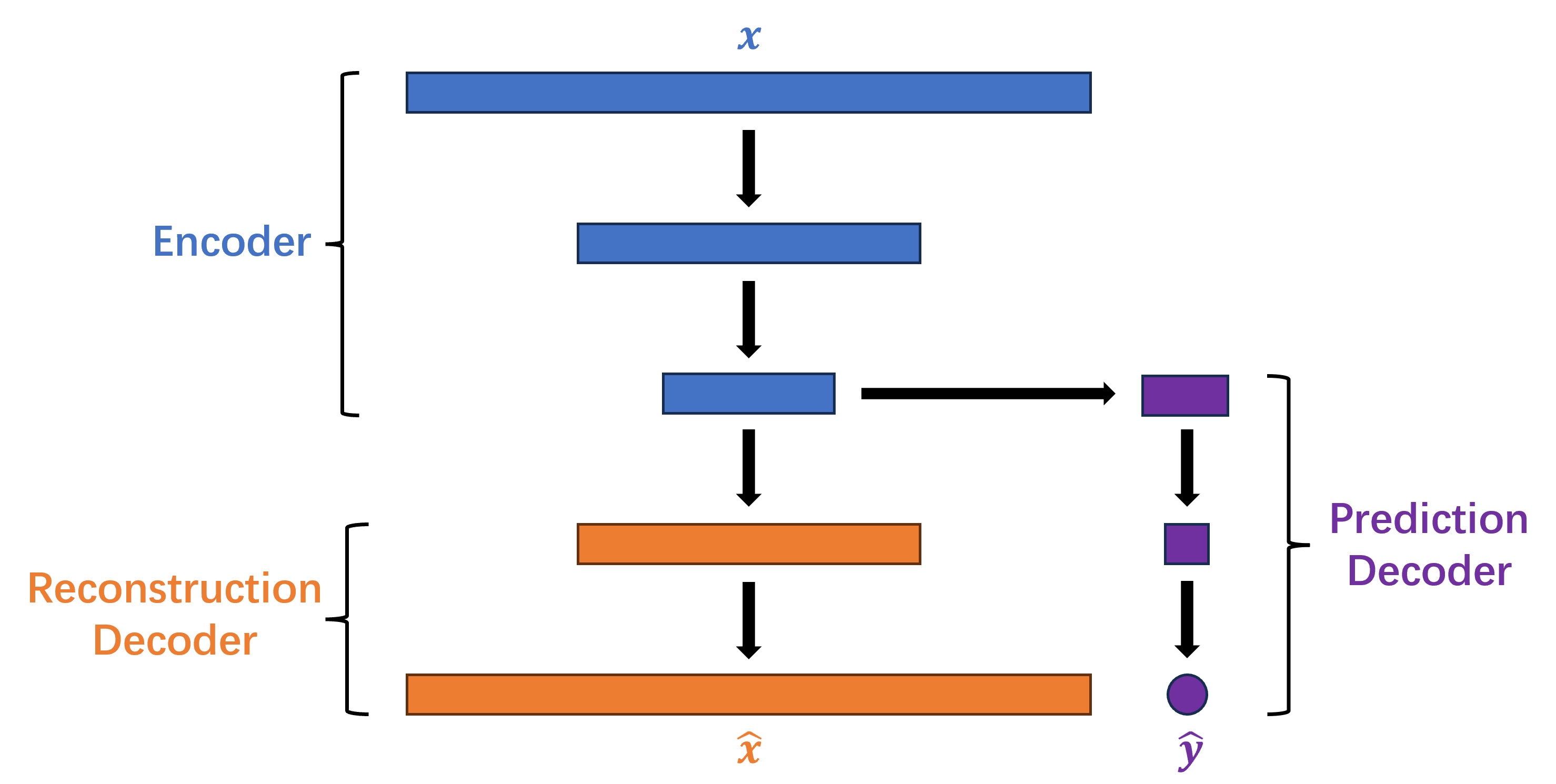}
    \caption{MEDN structure.}
    \label{fig:medn}
\end{figure}

A new regression model, namely MEDN (Multi-task Encoder-Decoder Network) is designed as an alternative solution for RM selection. Essentially, the model is composed of one encoder and two decoders, as shown in Figure \ref{fig:medn}. The encoder compresses the input features $x$ into the latent space and the prediction decoder utilizes the latent feature to produce the predicted inference latency as the output $\hat{y}$. Additionally, a reconstruction decoder is applied to rebuild the input features as its output $\hat{x}$, using the latent feature from the encoder. The reconstruction task aligns with the idea of autoencoders \cite{autoencoders}, which aim to formulate a more meaningful understanding of the input features. The training losses of the reconstruction task and prediction task are weighted sum into one single loss for the backpropagation procedure. 

For the experimental analysis in this paper, MLPs are applied as the underlying structure for both the encoder and the decoders. The structure of the reconstruction decoder is symmetric to that of the encoder, and the hidden dimensions in the prediction decoder gradually drop along the power of twos. Nevertheless, it is possible to specify a different MEDN configuration as long as the basic idea preserves the same. Theoretically, MEDN is more capable of capturing the input features and providing more accurate predictions, compared to simpler deep RMs like MLP. MEDN is also faster and lighter than Random Forest, whose inference speed and model size are easily affected by the size and complexity of the dataset.

\subsection{Auto-Selection}

Algorithm \ref{alg:auto-selection} illustrates how the framework manages to find an RM selection set automatically regarding the prediction accuracy and R-squared metric results. In addition, the algorithm examines an extra objective (time-per-sample/model-size) to further improve the performance in the time/space dimension. Specifically, the algorithm first filters a set of RMs possessing the highest prediction accuracy results (Line \ref{alg:p:filter-acc}-\ref{alg:p:tolerate-acc}). RMs with less than $\epsilon_a$ accuracy difference (as a tolerance factor) from the one with the highest accuracy are also included. Then, a similar filtering operation is performed on the R-squared metric (Line \ref{alg:p:filter-r2}--\ref{alg:p:tolerate-r2}). Finally, one RM is selected according to the specified extra objective (Line \ref{alg:p:extra-obj}), where the selected RM demonstrates the least inference time or model size. Utilizing filtering with tolerance, the Time/Space-efficient Auto-selection algorithm ensures that the overall accuracy and R-squared results are at the highest possible level, while attempting to place preferences on faster or lighter RMs. It is pertinent to note that when two RMs possess the same prediction accuracy, the one with a higher R-squared value is preferred. Intuitively, a higher R-squared value attempts to confirm that falsely predicted samples have smaller absolute errors, thus leading to a better result. Hence, the filtering operation is performed on accuracy first and R-squared afterwards.

\begin{algorithm}
\caption{Time/Space-efficient Auto-selection.}
\label{alg:auto-selection}
\begin{algorithmic}[1]
\REQUIRE Evaluation result set $Y$; accuracy and R-squared tolerance $\epsilon_a, \epsilon_r$
\ENSURE Selection $S$: module $m \Rightarrow$ regression model $u$
\STATE $o \gets$ time-per-sample (time-efficient) or model-size (space-efficient)
\STATE $S \gets \emptyset$
\FOR{each module $m \in M$}
    \STATE $Y_m \gets \{y \in Y \mid y.\text{m} = m\}$
    \STATE $y_1 \gets \arg \max\limits_{y \in Y_m} y.\text{acc}$ \label{alg:p:filter-acc}
    \STATE $Y_1 \gets \{y \in Y_m \mid y_1.\text{acc} - y.\text{acc} \le \epsilon_a \}$ \label{alg:p:tolerate-acc}
    \STATE $y_2 \gets \arg \max\limits_{y \in Y_1} y.\text{r}$ \label{alg:p:filter-r2}
    \STATE $Y_2 \gets \{y \in Y_1 \mid y_2.\text{r} - y.\text{r} \le \epsilon_r \}$ \label{alg:p:tolerate-r2}
    \STATE $y_s \gets \arg \min\limits_{y \in Y_2} y.\text{o}$ \label{alg:p:extra-obj}
    \STATE $S[m] \gets y_s.\text{u}$
\ENDFOR
\RETURN $S$
\end{algorithmic}
\end{algorithm}
\section{Experimental Evaluation}
\label{sec:exp}
\subsection{Configuration}

In total, four RMs including the proposed MEDN, Random Forest (abbreviated as RF), Multi-Layer Perceptron (abbreviated as MLP), as well as Linear Regression (abbreviated as LR) are trained and evaluated on nine modules, four of which are composite modules. The configurations of Random Forests and Multi-Layer Perceptrons are respectively the same as nn-Meter \cite{nn_meter} and SWFU \cite{swfu}. In total, 10,000 samples are generated for convolutional modules and 2,000 samples are generated for other modules. The datasets are split into train, validation, test set with the ratio of $7 \colon 1 \colon 2$. The input features are min-max scaled before being utilized by the RMs. All presented results are evaluated on test sets. To mimic a dynamic environment, random module inference jobs are occasionally generated on the device.


\setlength{\arrayrulewidth}{1pt}
\setlength{\tabcolsep}{5.0pt}
\renewcommand{\arraystretch}{1.1}
\begin{table}
\centering
\caption{Configurations of MEDN.}
\label{tab:medn-config}
\begin{tabular}{|c|c|c|}
\hline
\textbf{Module} & \textbf{Encoder Hidden} & \textbf{\makecell{Weight Ratio \\ (Reconstr : Pred)}} \\ \hline
avgpool & {[}64, 32, 16{]} & 0.001 \\
bn & {[}32, 16, 8{]} & 1.0 \\
conv & {[}128, 64, 32{]} & 100.0 \\
linear & {[}64, 32, 16{]} & 0.01 \\
maxpool & {[}64, 32, 16{]} & 0.01 \\ \hline
\end{tabular}
\end{table}


Table \ref{tab:medn-config} lists the configurations of MEDN, including the hidden dimensions of the encoder (excluding the input dimension) and the weight ratio which is exactly reconstruction weight divided by prediction weight. All deep RMs are trained for 500 epochs with the learning rate as $0.005$. All losses are calculated using the Smooth L1 Loss function. The repeated experiments are implemented with PyTorch \cite{pytorch} as well as scikit-learn \cite{sklearn} and conducted on Tesla P100 PCIe.
\subsection{Results on Single-RM Selection}

According to Table \ref{tab:res-rm}, other RMs like RF and MLP achieve lower overall prediction accuracy and R-squared compared to the proposed MEDN. Besides, they are generally less efficient both in time (time-per-sample as "Tps" in milliseconds) and space dimension (model-size as "Size" in kilobytes). The overall metric results are essentially percentage differences with MEDN, averaged over all modules. The results prove that MEDN serves better as an alternative RM solution on all evaluated metrics.

\setlength{\arrayrulewidth}{1pt}
\setlength{\tabcolsep}{1.26pt}
\renewcommand{\arraystretch}{1.1}
\begin{table}[ht]
\centering
\caption{Results on different regression models for multiple deep-learning modules.}
\label{tab:res-rm}
\begin{tabular}{|c|cccc|cccc|}
\hline
\multirow{2}{*}{\textbf{Module\textbackslash{}RM}} & \multicolumn{4}{c|}{\textbf{\underline{MEDN}}} & \multicolumn{4}{c|}{\textbf{RF}} \\ \cline{2-9} 
 & \multicolumn{1}{c|}{\textbf{Acc}} & \multicolumn{1}{c|}{\textbf{R$^2$}} & \multicolumn{1}{c|}{\textbf{Tps}} & \textbf{Size} & \multicolumn{1}{c|}{\textbf{Acc}} & \multicolumn{1}{c|}{\textbf{R$^2$}} & \multicolumn{1}{c|}{\textbf{Tps (ms)}} & \textbf{Size (KB)} \\ \hline
avgpool & \multicolumn{1}{c|}{0.6700} & \multicolumn{1}{c|}{0.9792} & \multicolumn{1}{c|}{0.4148} & 40 & \multicolumn{1}{c|}{0.7625} & \multicolumn{1}{c|}{0.9457} & \multicolumn{1}{c|}{83.4442} & 46089 \\
bn & \multicolumn{1}{c|}{0.8775} & \multicolumn{1}{c|}{0.9937} & \multicolumn{1}{c|}{0.8246} & 21 & \multicolumn{1}{c|}{0.9050} & \multicolumn{1}{c|}{0.9904} & \multicolumn{1}{c|}{50.4511} & 23595 \\
bn+relu & \multicolumn{1}{c|}{0.8300} & \multicolumn{1}{c|}{0.9750} & \multicolumn{1}{c|}{0.7895} & 21 & \multicolumn{1}{c|}{0.8300} & \multicolumn{1}{c|}{0.9934} & \multicolumn{1}{c|}{51.6820} & 23620 \\
conv+bn & \multicolumn{1}{c|}{0.6915} & \multicolumn{1}{c|}{0.7726} & \multicolumn{1}{c|}{0.4649} & 110 & \multicolumn{1}{c|}{0.5925} & \multicolumn{1}{c|}{0.7034} & \multicolumn{1}{c|}{114.5584} & 342353 \\
conv+bn+relu & \multicolumn{1}{c|}{0.6920} & \multicolumn{1}{c|}{0.7495} & \multicolumn{1}{c|}{0.6129} & 110 & \multicolumn{1}{c|}{0.5795} & \multicolumn{1}{c|}{0.6895} & \multicolumn{1}{c|}{147.0191} & 342261 \\
conv & \multicolumn{1}{c|}{0.6695} & \multicolumn{1}{c|}{0.8626} & \multicolumn{1}{c|}{0.4591} & 110 & \multicolumn{1}{c|}{0.5875} & \multicolumn{1}{c|}{0.7570} & \multicolumn{1}{c|}{116.1598} & 342490 \\
conv+relu & \multicolumn{1}{c|}{0.6925} & \multicolumn{1}{c|}{0.9618} & \multicolumn{1}{c|}{0.9540} & 110 & \multicolumn{1}{c|}{0.6165} & \multicolumn{1}{c|}{0.8489} & \multicolumn{1}{c|}{137.5373} & 342270 \\
linear & \multicolumn{1}{c|}{0.9600} & \multicolumn{1}{c|}{0.0328} & \multicolumn{1}{c|}{0.4048} & 40 & \multicolumn{1}{c|}{0.8850} & \multicolumn{1}{c|}{-1.6179} & \multicolumn{1}{c|}{76.6457} & 41028 \\
maxpool & \multicolumn{1}{c|}{0.6425} & \multicolumn{1}{c|}{0.8745} & \multicolumn{1}{c|}{0.4207} & 40 & \multicolumn{1}{c|}{0.7425} & \multicolumn{1}{c|}{0.7124} & \multicolumn{1}{c|}{83.5899} & 46100 \\ \hline
\textbf{\makecell{Avg diff\% \\ \textit{vs.} MEDN}} & \multicolumn{1}{c|}{\textbf{-}} & \multicolumn{1}{c|}{\textbf{-}} & \multicolumn{1}{c|}{\textbf{-}} & \textbf{-} & \multicolumn{1}{c|}{\textbf{-2.5\%}} & \multicolumn{1}{c|}{\textbf{-24\%}} & \multicolumn{1}{c|}{\textbf{+1.8e4\%}} & \textbf{+2.0e5\%} \\ \hline
\multirow{2}{*}{\textbf{Module\textbackslash{}RM}} & \multicolumn{4}{c|}{\textbf{MLP}} & \multicolumn{4}{c|}{\textbf{LR}} \\ \cline{2-9} 
 & \multicolumn{1}{c|}{\textbf{Acc}} & \multicolumn{1}{c|}{\textbf{R$^2$}} & \multicolumn{1}{c|}{\textbf{Tps}} & \textbf{Size} & \multicolumn{1}{c|}{\textbf{Acc}} & \multicolumn{1}{c|}{\textbf{R$^2$}} & \multicolumn{1}{c|}{\textbf{Tps (ms)}} & \textbf{Size (KB)} \\ \hline
avgpool & \multicolumn{1}{c|}{0.5600} & \multicolumn{1}{c|}{0.9035} & \multicolumn{1}{c|}{0.5509} & 77 & \multicolumn{1}{c|}{0.0950} & \multicolumn{1}{c|}{0.1603} & \multicolumn{1}{c|}{0.0732} & 3 \\
bn & \multicolumn{1}{c|}{0.8800} & \multicolumn{1}{c|}{0.9945} & \multicolumn{1}{c|}{0.6687} & 77 & \multicolumn{1}{c|}{0.2400} & \multicolumn{1}{c|}{0.7110} & \multicolumn{1}{c|}{0.1714} & 3 \\
bn+relu & \multicolumn{1}{c|}{0.8275} & \multicolumn{1}{c|}{0.9907} & \multicolumn{1}{c|}{0.4295} & 77 & \multicolumn{1}{c|}{0.2325} & \multicolumn{1}{c|}{0.3918} & \multicolumn{1}{c|}{0.1564} & 3 \\
conv+bn & \multicolumn{1}{c|}{0.5900} & \multicolumn{1}{c|}{0.7399} & \multicolumn{1}{c|}{1.0352} & 77 & \multicolumn{1}{c|}{0.1355} & \multicolumn{1}{c|}{0.0460} & \multicolumn{1}{c|}{0.0589} & 3 \\
conv+bn+relu & \multicolumn{1}{c|}{0.6370} & \multicolumn{1}{c|}{0.6304} & \multicolumn{1}{c|}{0.3112} & 77 & \multicolumn{1}{c|}{0.1545} & \multicolumn{1}{c|}{0.0342} & \multicolumn{1}{c|}{0.0924} & 3 \\
conv & \multicolumn{1}{c|}{0.4855} & \multicolumn{1}{c|}{0.8823} & \multicolumn{1}{c|}{0.3959} & 77 & \multicolumn{1}{c|}{0.1280} & \multicolumn{1}{c|}{0.0617} & \multicolumn{1}{c|}{0.1001} & 3 \\
conv+relu & \multicolumn{1}{c|}{0.5755} & \multicolumn{1}{c|}{0.9110} & \multicolumn{1}{c|}{0.3680} & 77 & \multicolumn{1}{c|}{0.1540} & \multicolumn{1}{c|}{0.0523} & \multicolumn{1}{c|}{0.1083} & 3 \\
linear & \multicolumn{1}{c|}{0.9550} & \multicolumn{1}{c|}{-0.0297} & \multicolumn{1}{c|}{0.3075} & 77 & \multicolumn{1}{c|}{0.9575} & \multicolumn{1}{c|}{0.0578} & \multicolumn{1}{c|}{0.0727} & 3 \\
maxpool & \multicolumn{1}{c|}{0.5775} & \multicolumn{1}{c|}{0.8937} & \multicolumn{1}{c|}{0.7146} & 77 & \multicolumn{1}{c|}{0.0875} & \multicolumn{1}{c|}{0.0393} & \multicolumn{1}{c|}{0.1477} & 3 \\ \hline
\textbf{\makecell{Avg diff\% \\ \textit{vs.} MEDN}} & \multicolumn{1}{c|}{\textbf{-7.1\%}} & \multicolumn{1}{c|}{\textbf{-3.2\%}} & \multicolumn{1}{c|}{\textbf{+1.4\%}} & \textbf{+77\%} & \multicolumn{1}{c|}{\textbf{-50\%}} & \multicolumn{1}{c|}{\textbf{-63\%}} & \multicolumn{1}{c|}{\textbf{-81\%}} & \textbf{-93\%} \\ \hline
\end{tabular}
\end{table}

With LR, only the Linear module is characterized by a high prediction accuracy, indicating that there is potentially a simple linear relationship between the input features and the output latency for this module. However, the R-squared results of RF and MLP turn negative on the Linear module. According to the detailed experiment result, the standard deviation of the dataset for this module is rather small. Therefore, R-squared results generally become negative when the Root Mean Squared Error (RMSE) values of RF ($0.0355$) and MLP ($0.021$) exceed the dataset standard deviation value ($0.0207$), while the proposed MEDN achieves a positive R-squared with an RMSE value of $0.0203$.

On the other hand, all RMs struggle with the latency prediction of convolutional modules due to the complicated module structures they possess. Nevertheless, MEDN manages to outperform other RMs by increasing the reconstruction task weight, as shown in Table \ref{tab:medn-config}. Besides, composite convolutional modules are generally easier to predict than the monolithic convolutional module, due to potential kernel fusion optimization techniques that are able to simplify the computation dynamics. Surprisingly, the prediction results on pooling layers are generally far from ideal. One of the reasons might be that these modules do not own weight parameters, thus lacking enough inferable information for deep RMs to comprehend. With a more complex model structure, especially like MEDN, this can easily lead to overfitting. Future research should consider adding more Inferable Parameters like the expected FLOPS (floating-point operations per second).
\subsection{Results on Ablation Experiment for MEDN}

To examine the effect of the reconstruction decoder in MEDN, an ablation experiment is performed to evaluate a non-reconstruction version (MEDN-Direct as MEDN-D) for comparison. Besides, MEDNs trained on different parameter sets are evaluated. MEDN-No-Infer (MEDN-XI) and MEDN-No-Measure (MEDN-XM) handle all parameters except the Inferable/Measurable Parameters. MEDN-RAW (MEDN-R) handles only Sampling Parameters.

\begin{figure}
    \centering
    \includegraphics[width=0.9\textwidth]{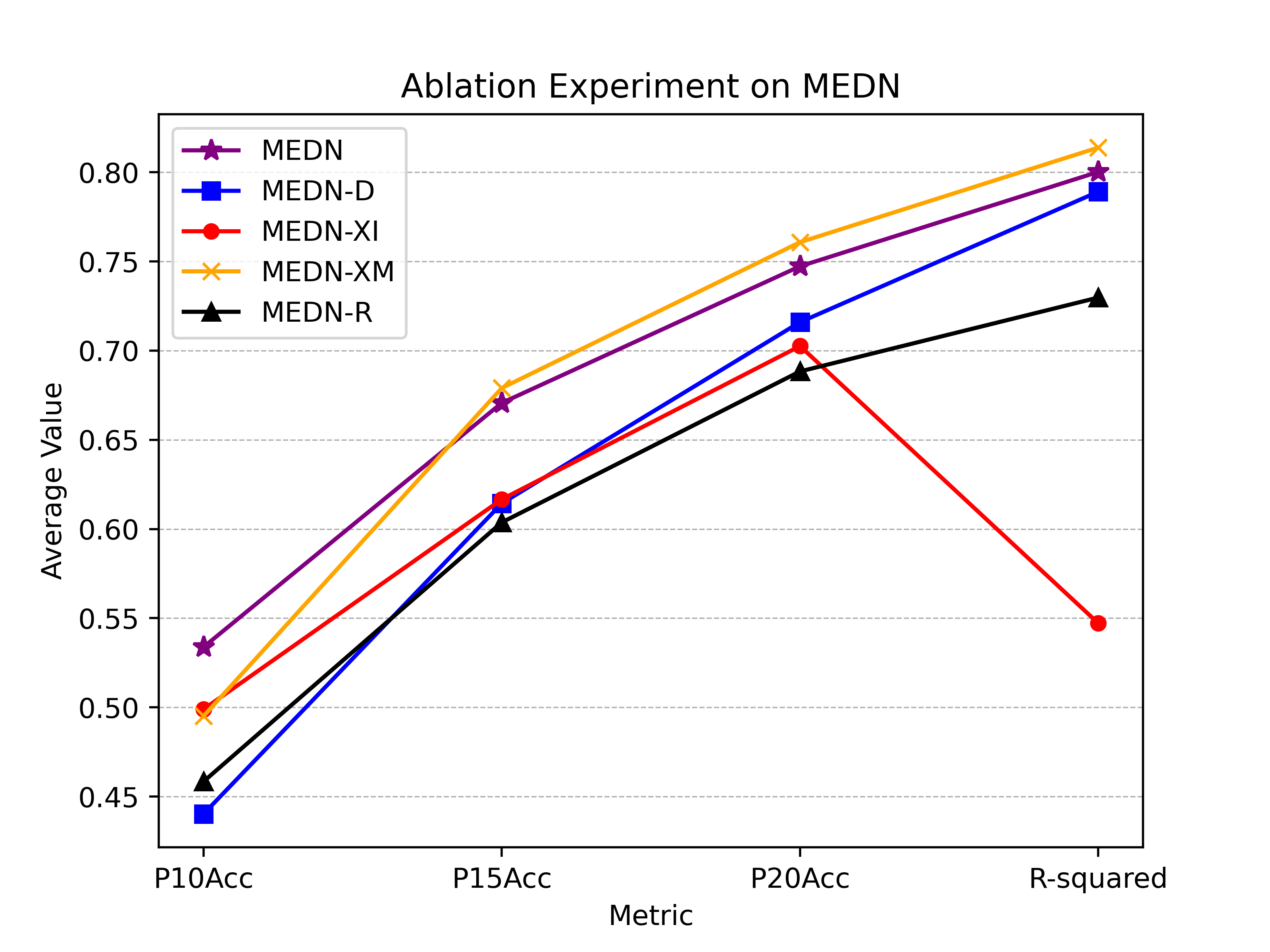}
    \caption{Ablation experiment on MEDN.}
    \label{fig:ablation}
\end{figure}

Figure \ref{fig:ablation} shows the evaluated metric results averaged over all modules. P20-Accuracy is the default accuracy metric in other result tables, specifying the ratio of samples whose prediction percentage error is within 20\%. First, it can be noticed that MEDN outperforms MEDN-D on all metrics, verifying that the reconstruction decoder serves an important role in the improved performance. Next, MEDN-XI performs generally worse than MEDN-XM but better than MEDN-R, indicating that Inferable Parameters provide more helpful guidance for latency prediction than Measurable Parameters, although both lead to better accuracy results. Finally, it is discovered that Measurable Parameters allow MEDN to be more accurate on the correctly predicted samples, resulting in higher P10-Accuracy than that of MEDN-XM.
\subsection{Results on Auto-Selection}

Table \ref{tab:res-alg} records the resulted RM selection sets of the Time-efficient Auto-selection algorithm (time-per-sample as the extra objective) and Space-efficient Auto-selection algorithm (model-size as the extra objective). Both algorithms set $\epsilon_a = \epsilon_r = 0.05$, specifying the tolerance as 5\%. Compared to the MEDN single-selection scheme where MEDN is selected for all modules, both selection sets manage to achieve a higher overall prediction accuracy and R-squared value. In detail, MEDN variants with the highest P20-Accuracy results are selected for the convolutional modules, and RF is selected for the max pooling layer where MEDN does not perform ideally. LR is selected for the simple Linear module, which results in a significant decrease in the prediction time and storage space.

"Results on diff" in the table record the original metric results of the RMs that differ among the schemes. Since both algorithms prioritize accuracy, the selected RMs for the modules are rather similar except for one difference on the composite Batch Normalization + ReLU module. Examining the difference, it can be discovered that for the Time-efficient Auto-selection algorithm, the inference time per sample of the MLP on the module is lower than that of MEDN-XM by the Space-efficient Auto-selection algorithm, while the model size is larger. This aligns with their extra objectives to save time and space respectively.

\setlength{\arrayrulewidth}{1pt}
\setlength{\tabcolsep}{5.0pt}
\renewcommand{\arraystretch}{1.1}
\begin{table}
\centering
\caption{Results on auto-selection schemes compared to MEDN single-selection.}
\label{tab:res-alg}
\begin{tabular}{|cc|c|c|}
\hline
\multicolumn{2}{|c|}{\textbf{Module/Scheme}} & \textbf{\begin{tabular}[c]{@{}c@{}}Time-efficient\\ Auto-selection\end{tabular}} & \textbf{\begin{tabular}[c]{@{}c@{}}Space-efficient\\ Auto-selection\end{tabular}} \\ \hline
\multicolumn{2}{|c|}{avgpool} & MEDN-XM & MEDN-XM \\
\multicolumn{2}{|c|}{bn} & MEDN-XM & MEDN-XM \\
\multicolumn{2}{|c|}{bn+relu} & {\ul MLP} & {\ul MEDN-XM} \\
\multicolumn{2}{|c|}{conv+bn} & MEDN-R & MEDN-R \\
\multicolumn{2}{|c|}{conv+bn+relu} & MEDN-XM & MEDN-XM \\
\multicolumn{2}{|c|}{conv} & MEDN-XM & MEDN-XM \\
\multicolumn{2}{|c|}{conv+relu} & MEDN-XM & MEDN-XM \\
\multicolumn{2}{|c|}{linear} & LR & LR \\
\multicolumn{2}{|c|}{maxpool} & RF & RF \\ \hline
\multicolumn{1}{|c|}{\multirow{2}{*}{{\textbf{\makecell{Results \\ on \underline{diff}}}}}} & \textbf{Tps (ms)} & \textbf{0.4295} & \textbf{0.4820} \\
\multicolumn{1}{|c|}{} & \textbf{Size (KB)} & \textbf{77} & \textbf{20} \\ \hline
\multicolumn{1}{|c|}{\multirow{2}{*}{\textbf{\makecell{Avg diff\% \\ \textit{vs.} MEDN}}}} & \textbf{Acc} & \textbf{+2.4\%} & \textbf{+2.63\%} \\
\multicolumn{1}{|c|}{} & \textbf{R$^2$} & \textbf{+0.41\%} & \textbf{+0.36\%} \\ \hline
\end{tabular}
\end{table}

\section{Conclusion}
\label{sec:conclusion}

In this paper, a flexible accuracy-oriented DNN module inference latency prediction framework is proposed to support customizable parameter configurations and accuracy-oriented RM auto-selection in a dynamic environment. A newly designed RM, namely MEDN, is also proposed as an alternative RM solution. The quantitative analysis presented in this paper demonstrates that MEDN is fast and lightweight, and capable of achieving the highest overall prediction accuracy and R-squared value. The Time/Space-efficient Auto-selection algorithm also manages to improve the overall accuracy by 2.5\% and R-squared by 0.39\%, compared to the MEDN single-selection scheme. In the future, more Inferable Parameters and Measurable Parameters will be utilised in the experimental analysis in order to fully exploit the feature comprehension ability of MEDN.
\subsubsection{Acknowledgements} This research was supported by: Shenzhen Science and Technology Program,  China (No. GJHZ20210705141807022); Guangdong Province Innovative and Entrepreneurial Team Programme, China (No. 2017ZT07X386); SUSTech Research Institute for Trustworthy Autonomous Systems, China. Corresponding author: Georgios Theodoropoulos.
%
%
\bibliographystyle{splncs04}
\bibliography{refs}

\end{document}